\documentclass[runningheads]{llncs}
\usepackage[T1]{fontenc}
\usepackage{booktabs}
\usepackage{graphicx}
\usepackage{todonotes}
\usepackage{subcaption}
\usepackage{listings}
\usepackage{multirow}
\usepackage{amssymb}
\usepackage[inline]{enumitem}

\begin{document}
\title{IDEA2: Expert-in-the-loop competency question elicitation for collaborative ontology engineering}
%
\titlerunning{IDEA2: Expert-in-the-loop competency question elicitation}
%
\author{
Elliott Watkiss-Leek\inst{1}\orcidID{0009-0002-9284-6590} \and
Reham Alharbi\inst{2}\orcidID{0000-0002-8332-3803} \and
Harry Rostron\inst{3} \and
Andrew Ng\inst{3} \and
Ewan Johnson\inst{3} \and
Andrew Mitchell\inst{3} \and
Terry R. Payne\inst{1}\orcidID{0000-0002-0106-8731} \and
Valentina Tamma\inst{1}\orcidID{0000-0002-1320-610X} \and
Jacopo de Berardinis\inst{1}\orcidID{0000-0001-6770-1969}
}

\authorrunning{E. Watkiss-Leek et al.}
%
\institute{
School of Computer Science and Informatics, University of Liverpool, Brownlow Hill, Liverpool, L69 7ZX, UK \\
\email{\{e.watkiss-Leek,jacodb\}@liverpool.ac.uk}
\and
College of Computer Science and Engineering, Taibah University, Saudi Arabia \\
\email{rfalharbi@taibahu.edu.sa}
\and
Unilever Plc., Materials Innovation Factory, University of Liverpool, 51 Oxford Street, Liverpool, L7 3NY, UK
}
\maketitle              
\begin{abstract}
Competency question (CQ) elicitation represents a critical but resource-intensive bottleneck in ontology engineering.
This foundational phase is often hampered by the communication gap between domain experts, who possess the necessary knowledge, and ontology engineers, who formalise it.
This paper introduces IDEA2, a novel, semi-automated workflow that integrates Large Language Models (LLMs) within a collaborative, expert-in-the-loop process to address this challenge.
The methodology is characterised by a core iterative loop: an initial LLM-based extraction of CQs from requirement documents, a co-creational review and feedback phase by domain experts on an accessible collaborative platform, and an iterative, feedback-driven reformulation of rejected CQs by an LLM until consensus is achieved.
To ensure transparency and reproducibility, the entire lifecycle of each CQ is tracked using a provenance model that captures the full lineage of edits, anonymised feedback, and generation parameters.
The workflow was validated in 2 real-world scenarios (scientific data, cultural heritage), demonstrating that IDEA2 can accelerate the requirements engineering process, improve the acceptance and relevance of the resulting CQs, and exhibit high usability and effectiveness among domain experts.
We release all code and experiments at \texttt{\url{https://github.com/KE-UniLiv/IDEA2}}.
\end{abstract}

\section{Introduction}\label{sec:introduction}

Ontologies serve as the formal, explicit specification of shared conceptualisations, enabling interoperable knowledge sharing across heterogeneous information systems \cite{studer1998knowledge}.
The process of creating these artefacts, known as Ontology Engineering (OE), is a complex, time-consuming, and often manual endeavour that demands a synergistic combination of skills in knowledge representation, logic, and deep domain expertise \cite{kendall2019ontology}.
Traditional approaches to requirements engineering in OE involve eliciting requirements through interviews, workshops, or co-production sessions with stakeholders \cite{malone2014software}.
These specifications are then reorganised into user stories or use case definitions \cite{kendall2019ontology}, from which competency questions (CQs) \cite{GruningerFox94} are derived.
CQs provide a set of natural language questions that an ontology should satisfy, and underpin ontology design in established methodologies such as eXtreme Design \cite{presutti2009extreme}, NeOn \cite{suarez2011neon}, and Methontology \cite{fernandez1997methontology}.

Despite their importance, the process of authoring high-quality CQs is notoriously difficult \cite{monfardini2023}.
This is a significant bottleneck within the OE lifecycle, as it requires extensive interaction between knowledge engineers and domain experts (DE), a process that can be inefficient and prone to ambiguity \cite{deberardinis2023polifonia}.
The collection of ontology requirements and the extraction of CQs from textual sources are perceived as among the most challenging and time-consuming tasks in the field \cite{monfardini2023,zhang2024ontochat}.
This creates a knowledge gap: \textit{ontology engineers} possess the skill to formalise verifiable and possibly atomic CQs, but lack specific domain expertise; 
 conversely, \textit{domain experts} possess deep scientific knowledge but often struggle to formulate requirements as formal CQs.
This highlights a critical need for methodologies and tools that can streamline this process, reduce manual effort, and bridge the gap between domain experts and ontology engineers.

The recent proliferation of Large Language Models (LLMs) has introduced a transformative potential across numerous domains, including software and knowledge engineering \cite{zhang2023using,marques2025enhancing}.
With their advanced capabilities in natural language understanding and generation, LLMs are increasingly being applied to automate and assist with complex, language-intensive requirements engineering tasks \cite{zadenoori2025large}.
These models can process vast amounts of unstructured text, e.g. user stories, technical manuals, and interview transcripts, to assist in the elicitation, analysis, and validation of requirements, thereby reducing the complexity and human effort traditionally involved \cite{marques2025enhancing,zadenoori2025large}.

However, while LLMs can assist in the automation of the CQ generation process, their application in domain-specific contexts is not without challenges \cite{alharbi2025comparative}.
Known limitations such as factual inaccuracies (``hallucinations''), a lack of deep domain grounding, and the potential for generating plausible but incorrect information necessitate a cautious approach \cite{garijo2024large,xi2025rise}.
A fully automated pipeline risks producing a set of CQs that are syntactically valid but semantically misaligned with the nuanced requirements of the domain experts.

To overcome these issues, we introduce IDEA2, a semi-automated, expert-in-the-loop workflow that positions the LLM as a requirement elicitator within a human-centric, collaborative process \cite{zhang2025towards}.
The core idea of IDEA2 is to separate requirement generation from their validation, therefore creating a clear division of tasks that improves the efficiency of requirements engineering.

The workflow establishes a structured feedback loop between the LLM (acting as the requirement elicitation agent) and domain experts (DE, acting as validators), mediated by an accessible collaborative platform where feedback is collected and discussions can unfold in a structured manner.
This process begins with an LLM performing an initial extraction of candidate CQs from multiple sources (e.g., personas, user stories, legacy schema definitions).
These CQs are then presented to DEs for a co-creational review, where the DEs provide structured feedback (ratings, tags, comments).
This feedback, in turn, is used to programmatically guide a second LLM instance to iteratively reformulate and refine the CQs until a stopping criterion is met (e.g., number of iterations, experts' consensus).
This process ensures that the final set of requirements is computationally derived while being rigorously validated by those with the deepest understanding of the domain.
We validate the proposed workflow in a domain-intensive scenario focusing on scientific data standards, involving four domain experts from our industry partner, Unilever PLC.
Furthermore, we assess its reformulation capabilities in a second real-world scenario within the cultural heritage domain, specifically targeting collection management.
Our results demonstrate that IDEA2 accelerates the requirement engineering process and improves the acceptance and relevance of the resulting CQs, while exhibiting high usability and effectiveness among domain experts. This paper makes the following contributions:
\begin{itemize}
    \item \textbf{A human-in-the-loop workflow (IDEA2)} for CQ engineering that synergistically combines LLM-based generation with structured, collaborative feedback from domain experts for iterative refinement.
    \item \textbf{A collaborative dashboard integration} that leverages a widely accessible, non-specialist platform (Notion) via its API to lower the barrier to adoption for interdisciplinary teams and facilitate collaboration.
\end{itemize}
We validate the workflow empirically across two real-world domains, demonstrating its practical applicability and effectiveness.
The remainder of the paper is organised as follows: Section~\ref{sec:related} reviews work in LLM-based OE requirements elicitation;
Section~\ref{sec:method} details IDEA2 and its collaborative workflow.
Section~\ref{sec:experiments} and \ref{sec:discussion} report on the experimental validation across scientific and cultural heritage domains, discussing the results and limitations.
\section{Related work}\label{sec:related}

The field of OE has evolved from waterfall-style methodologies to more flexible, iterative approaches.
Early frameworks such as METHONTOLOGY provided a structured, step-by-step process covering the entire ontology lifecycle, from specification to maintenance \cite{fernandezlopez1997methontology}.
While foundational, these methods were often rigid and resource-intensive.
In response, the community has increasingly adopted agile principles to support iterative development.
Methodologies like Neon \cite{suarez2011neon} or eXtreme Design (XD) foster the incremental development of ontologies by addressing small sets of requirements, often formalised as user stories and CQs, and encourage the reuse of ontology design patterns \cite{blomqvist2010experimenting}.

The application of LLMs to OE is a rapidly expanding area of research.
Garijo et al. \cite{garijo2024large} provide an initial mapping of this space, identifying the diverse roles that LLMs can assume within the OE process, including acting as surrogate domain expert, ontology engineer's assistant, or automated evaluator.
Research in this area is broad, covering tasks from ontology matching and alignment \cite{xi2025rise} to the more foundational task of ontology learning, where recent work investigates the use of LLMs for core ontology construction tasks such as term typing, taxonomy discovery, and non-taxonomic relation extraction from text~\cite{babaeigiglou2023llms4ol,llms4ol2024challenge,lippolis2025ontogenia}.

A growing body of work has focussed on the automated generation of CQs. The Retrofit-CQ approach, for example, addresses the common problem of ontologies being published without their corresponding CQs \cite{alharbi2024investigating}.
Its pipeline reverse-engineers candidate CQs by extracting triples from an \textit{existing} ontology and using an LLM to generate questions that these triples could answer \cite{alharbi2023experiment}.
Similarly, the RevOnt approach generates CQs by reverse-engineering them from a populated knowledge graph (KG), verbalising KG triples and transforming them into questions \cite{ciroku2024revont,ciroku2024revont}.
Other methods \cite{pan2024rag} extract CQs from scientific papers through a retrieval-augmented generation (RAG) approach or explore CQ extraction from KGs constructed from textual data via LLMs \cite{nuzzo2024automated}.
While these methods are valuable for the post-hoc documentation of existing semantic artefacts, IDEA2 addresses the \textit{ab initio} elicitation of requirements for new projects, operating on sources \textit{before} an ontology has been constructed.

The recognition of LLM limitations, such as their propensity for factual hallucination, has underscored the necessity of human-in-the-loop paradigms (HITL) \cite{garijo2024large,xi2025rise}.
The IDEA2 workflow is built on this principle, ensuring that domain experts are the ultimate arbiters of quality. IDEA2 evolves and extends the \textbf{IDEA} framework, that first introduced NLP methods to support the CQ lifecycle and to assist knowledge engineers by 1) automatically tagging CQs with potential issues (e.g., 'complex', 'nested') and 2) 
clustering similar requirements for manual analysis \cite{deberardinis2023polifonia}.
In contrast, IDEA2 takes a more prominent human-in-the-loop approach through the automatic extraction and refinement of CQs by means of an iterative, feedback-driven reformulation loop.

Another relevant work is \textbf{OntoChat}, a conversational framework for OE that uses an LLM-based agent to guide a user through the creation of user stories and the extraction of CQs \cite{zhang2024ontochat}.
IDEA2 shares OntoChat's goal of LLM-assisted requirement elicitation, but it employs an asynchronous, conversational-based model.
This architectural choice is inherently more scalable for projects involving large or geographically distributed teams of domain experts, allowing numerous experts to contribute feedback independently, which the system can then programmatically aggregate for the reformulation step.


A summary of the key differences between IDEA2 and existing approaches for competency question generation is outlined in Table~\ref{tab:cq_comparison} for reference.

\begin{table}[b]
\centering
\begin{tabular}{lccc}
\hline
Approach & CQ Generation & Expert Validation & Iterative Reformulation \\
\hline
IDEA \cite{deberardinis2023polifonia} & $\times$ & $\checkmark$ & $\times$ \\
Retrofit-CQ \cite{alharbi2023experiment} & $\checkmark$ & $\times$ & $\times$ \\
RevOnt \cite{ciroku2024revont} & $\checkmark$ & $\times$ & $\times$ \\
OntoChat \cite{zhang2024ontochat} & $\checkmark$ & $\checkmark$ & $\times$ \\ \hline
\textbf{IDEA2} & $\checkmark$ & $\checkmark$ & $\checkmark$ \\
\hline
\end{tabular}
\caption{Comparison of approaches for competency question elicitation.}
\label{tab:cq_comparison}
\end{table}


\section{The IDEA2 workflow}\label{sec:method}

We address the knowledge acquisition bottleneck in OE by introducing IDEA2, a collaborative methodology designed to iteratively extract, review, and reformulate Competency Questions (CQs) using Large Language Models (LLMs) with expert-in-the-loop validation.
The central premise of the framework is to decouple requirement elicitation, delegated to the LLM, from requirement validation, reserved for domain experts. 
This separation of concerns maximises the efficiency of the requirements engineering process by allowing experts to focus on high-value verification tasks rather than the manual authoring of requirements.

The system extracts batches of CQs from heterogeneous requirement specifications, including narrative sources (e.g., personas, user stories) and structured schema definitions.
This flexibility enables the workflow to support both requirement elicitation and retrofitting with minimal OE supervision
Each CQ is tracked with its generation provenance (configuration, iteration number, version history) and presented in a validation dashboard for expert review.
This ensures that the final requirements accurately reflect the semantics of the source scenario as understood by domain stakeholders, while maintaining sufficient abstraction to address the expected system functionalities.

\begin{figure}[t]
    \centering
    \includegraphics[width=\textwidth, trim={22pt} 0 {20pt} 0, clip]{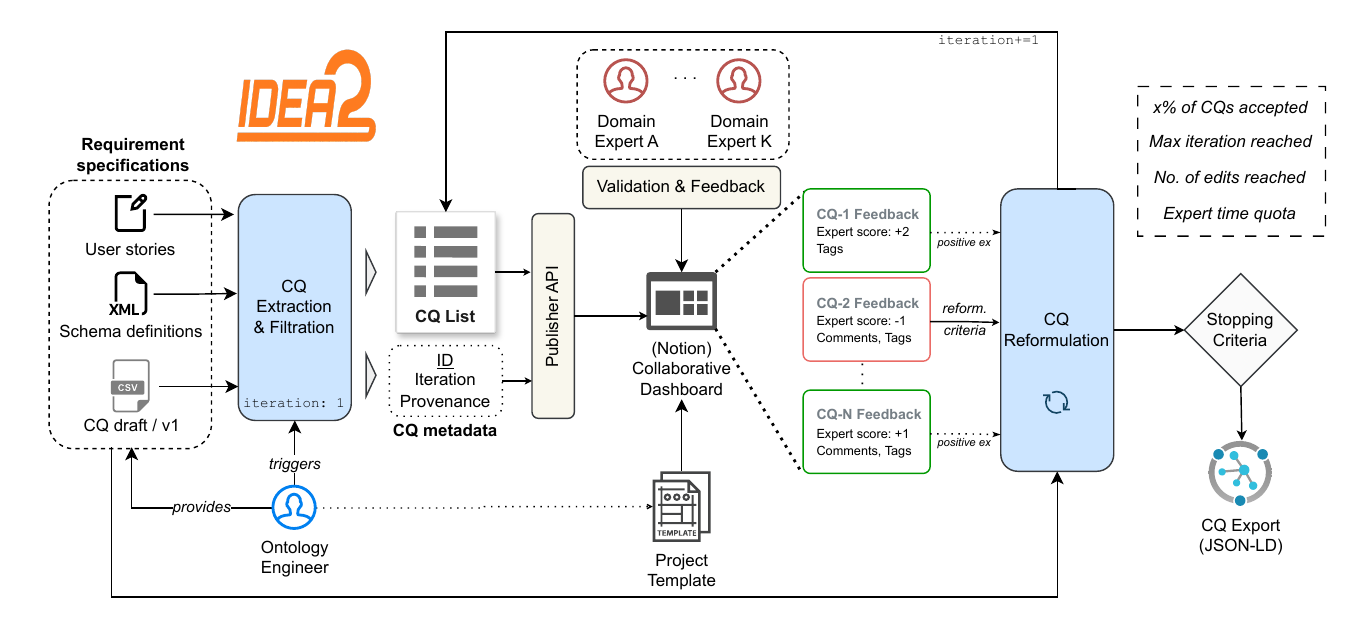}
    \caption{Overview of the IDEA2 architecture: (1) \emph{CQ Extraction} generates candidate requirements from heterogeneous sources; (2) these are published to a collaborative dashboard for \emph{Expert Validation}; (3) feedback drives an \emph{Iterative Reformulation} loop where rejected CQs are refined by the LLM based on the feedback; and (4) the process terminates upon meeting defined stopping criteria, producing a CQ export.}
    \label{fig:arch}
\end{figure}

The architecture of IDEA2 is illustrated in Figure~\ref{fig:arch}.
The workflow is a structured four-phase process designed to guide a team from initial unstructured source materials to a validated set of CQs.
The workflow is cyclical, with artefacts flowing between an LLM and a shared collaborative workspace where domain experts interact.
The process is composed of four primary phases: (1) multi-source CQ Extraction, (2) collaborative expert-in-the-loop validation and refinement, (3) iterative CQ reformulation, and (4) continuous provenance tracking and publishing.
Each of these steps is detailed in the following subsections.

\begin{figure}[t]
    \centering
    \includegraphics[width=\textwidth]{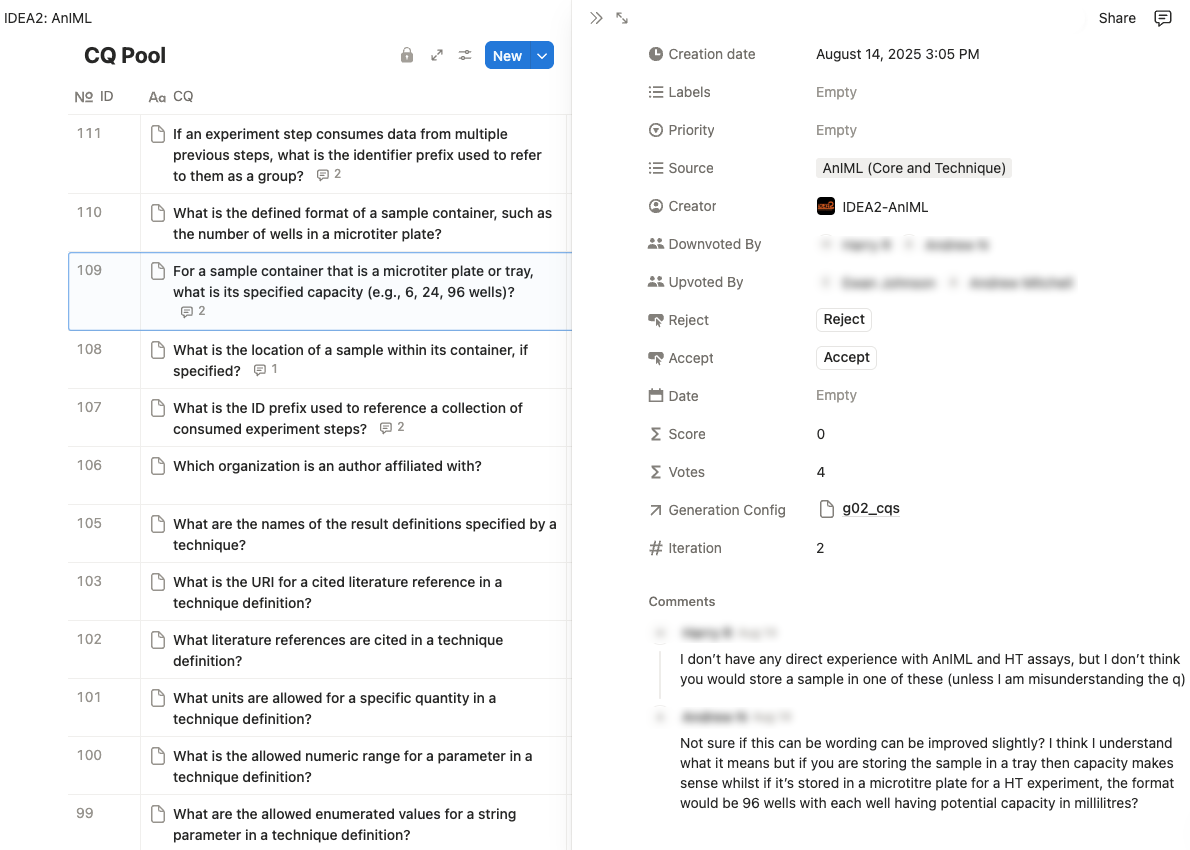}
    \caption{Screenshot of the IDEA2 collaborative dashboard on Notion. The interface shows the \textit{CQ Pool} (left) and the detailed validation view (right) for a specific requirement (ID 109). In this example, a split vote (Score 0) and qualitative comments from domain experts highlight a domain-specific ambiguity, providing the necessary context to trigger and guide the reformulation loop.}
    \label{fig:notion}
\end{figure}

\subsection{Multi-source CQ extraction}

The workflow initiates with the ingestion of heterogeneous source documents, such as user stories, use case descriptions, or XML schema definitions.
User stories are narrative-driven, focusing on the ``who'' (persona) and the ``why'' (goal/benefit), often leaving specific system interactions implicit \cite{carriero2019arco}.
Similarly, use cases provide a structured, step-by-step description of system behaviour, explicitly detailing the ``how'' through actors, preconditions, and flow of events \cite{kendall2019ontology}.
In contrast, XML schemas are widely used to define the syntactic structure of data for exchange, and deriving ontologies from their Document Type Definition is an emerging direction \cite{Duan2023_XSD2SHACL}.

The core of this phase is a prompt-driven extraction process, where an LLM is guided by an engineered prompt to identify and formulate potential CQs.
To ensure the generation of high-quality requirements, the \emph{CQ formulation prompt} adapts the optimisation strategy proposed in OntoChat \cite{zhang2024ontochat}.
The extraction pipeline employs a few-shot learning approach that executes a multi-step refinement chain to enforce two key properties: \emph{atomicity} and \emph{abstraction}.

First, the model detects and decomposes complex questions into atomic units.
For instance, if a source implies multiple requirements (e.g., ``\textit{What genres and styles are associated with Penny Lane?}''), the system splits this into distinct CQs for ``genres'' and ``styles'' respectively.
Second, the pipeline enforces named entity abstraction to ensure the requirements are defined at the conceptual level rather than the instance level.
Guided by examples, the LLM generalises specific references (e.g., replacing the entity ``Penny Lane'' with the concept ``musical work''), ensuring the resulting CQ targets concepts rather than A-Box data.

While the process is primarily automated, the workflow remains hybrid; domain experts can manually author or inject CQs into the pool at any stage.
Finally, the set of candidate CQs is filtered using sentence embedding similarity to remove duplicates \cite{ciroku2024revont} before being passed to the validation phase.

\subsection{Collaborative feedback collection}
The filtered CQs are systematically published via API to a collaborative dashboard implemented on Notion\footnote{\url{http://notion.so}}.
This architectural choice replaces bespoke, specialized OE tools with a widely accessible platform, significantly lowering the barrier to entry for domain experts—such as scientists, curators, or scholars—who may lack software engineering training \cite{garijo2024large}.
Within this workspace, each CQ is represented as a structured database entry, allowing for the capture of machine-readable feedback and provenance data.

During each iteration, annotators review the candidate CQs and are instructed to make a binary decision to \textit{accept} or \textit{reject} each CQ.
A CQ may be rejected based on the following criteria:
\begin{itemize}
    \item \textbf{Relevance and consistency:} The CQ is not entailed by the source text (hallucination) or explicitly contradicts the user stories or personas.
    \item \textbf{Abstraction level:} The CQ contains lower-level implementation details (e.g., specific XML schema element names, camelCasing) rather than conceptual entities.
    \item \textbf{Formulation:} The CQ is ambiguous, linguistically unclear, or lacks atomicity (i.e., it entails multiple distinct requirements).
    \item \textbf{Refinement failure:} In later iterations, a CQ may be rejected if it fails to adequately address the specific feedback provided in a previous loop.
\end{itemize}

The annotation platform supports a three-tiered feedback mechanism to guide the reformulation process:
\begin{itemize}
    \item \textbf{Voting.} Experts cast an ``accept'' or ``reject'' vote. These are aggregated into a numerical consensus score which determines if the CQ proceeds to the ontology or returns to the LLM for refinement.
    \item \textbf{Structured feedback.} Experts can apply thematic tags to categorize requirements. Furthermore, we implement a priority labelling mechanism inspired by the ``Buy a Feature'' method \cite{kirk2011democracy}. This allows the collaborative ranking of requirements, enabling stakeholders to explicitly bid on the features most critical to their domain needs.
    \item \textbf{Qualitative comments.} Experts provide free-text comments justifying rejections. These comments are crucial, as they are injected into the LLM context window to steer the subsequent reformulation.
\end{itemize}

\paragraph{\textbf{Dashboard implementation}.}
A screenshot of the dashboard used for the evaluation (c.f. Section~\ref{sec:experiments}) is provided in Figure~\ref{fig:notion}.
While the familiar interface benefits the user, the decision to leverage the Notion API was primarily driven by the need for rapid prototyping and deployment.
Unlike bespoke OE tools that often require complex installation or local configuration, this cloud-native approach eliminates technical overhead, allowing the workflow to be instantiated immediately.
To further accelerate adoption and ensure reproducibility, we release a pre-configured IDEA2 project template and webhook integration, simplifying the setup process and facilitating the reuse of the workflow across different domains.

\subsection{Iterative CQ reformulation}

The reformulation phase acts as a quality control loop, ensuring that no requirement enters the ontology design specification without explicit validation.
We employ a consensus-based mechanism where any CQ failing to achieve a positive net score ($\leq 0$) is automatically queued for revision \cite{deberardinis2023polifonia}.
While the default implementation uses a simple majority vote, the architecture can be extended with pluggable consensus models -- such as weighted voting based on self-reported expertise or different strategies \cite{sheshadri2013square}.
This validation cycle offers two primary benefits: (1) it acts as a semantic filter, allowing domain experts to intercept LLM hallucinations or irrelevant requirements before they propagate; and (2) it creates a feedback-driven loop that steers the model to refine requirements based on specific human critique.

To execute a revision, the system reuses a highly contextualised \textit{reformulation prompt}.
In contrast to the initial extraction prompt (which relies primarily on few-shot examples), the reformulation prompt injects the full provenance history of the requirement.
This context window includes the original requirement specifications (user stories, schemas), the initial CQ formulation, and the complete validation log—comprising scores, votes, and, crucially, the anonymised qualitative comments from experts.
This allows the LLM to ``reason'' over the rejection; for example, if an expert rejects a CQ due to ambiguity, the model uses the comment to disambiguate the specific term in the next draft.
The reformulated CQs are then re-published to the dashboard, triggering a new validation cycle.

The iterative process continues until a predefined stopping criterion is met, allowing teams to balance requirement quality against resource constraints.
The specific criterion depends on the project needs; examples include:
\begin{itemize}
    \item \textbf{Convergence threshold:} the process terminates when a target percentage of CQs (e.g., 90\%) reaches a positive consensus state.
    \item \textbf{Iteration cap:} a hard limit on the number of feedback loops to prevent infinite refinement cycles on contentious requirements.
    \item \textbf{Modification stability:} the process stops if the semantic distance between reformulations drops below a threshold.
    \item \textbf{Resource quota:} a limit based on expert time availability or token usage.
\end{itemize}

\paragraph{\textbf{Implementation}.}
The workflow leverages \texttt{Gemini 2.5 Pro} via the Google Cloud API.
This specific model was selected for its extensive context window (up to 1M tokens), enabling the ingestion of large-scale requirement specifications and the maintenance of the full iteration backlog.
To perform effective reformulation, the model must simultaneously hold the entire requirement specification (often comprising lengthy schemas or multimedia user stories) alongside the cumulative history of generated CQs and expert feedback.
While the current implementation is optimised for large-context models, the underlying workflow is agnostic and allows interfacing with other LLMs via their API.

\subsection{Provenance and versioning}
A comprehensive provenance record is maintained for each CQ.
This record encapsulates essential metadata, including a unique identifier, the requirement's origin (distinguishing between LLM-extracted and human-authored entries), and a direct link to the source specification document.
Crucially, the record is linked to a generation configuration object that captures the precise parameters used during elicitation -- such as the model checkpoint, seed, temperature, TopP, and frequency penalty -- allowing for the exact replication of the generation process.
The provenance model also tracks the full lifecycle of the requirement, maintaining a version history that links original CQs to their reformulations and logs the detailed outcomes of each review cycle.

The workflow supports the export of generated CQs into a standardised JSON-LD format.
This export schema integrates multiple established vocabularies to maximise interoperability: it reuses the CQ definition from the OWLUnit ontology\footnote{\texttt{\url{https://github.com/luigi-asprino/owl-unit}}}, employs PROV-O \cite{moreau2011open} as the core lineage model, and structures the dataset according to the Croissant data model \cite{akhtar2024croissant}.
Overall, this export feature aims to facilitate the reuse of requirements for ontology design and also to incentivise the formal publication of CQ datasets -- a practice that, despite its value, remains under-adopted by the community \cite{alharbi2024role,potoniec2020dataset}.
\section{Experiments}\label{sec:experiments}

To validate the efficacy of the proposed methodology, we evaluate IDEA2 across two distinct application scenarios designed to measure its effectiveness, efficiency, and usability.
To ensure reproducibility, all code, experimental data, and the configuration parameters used in these studies are openly available at \texttt{\url{https://github.com/KE-UniLiv/IDEA2}}. 

\subsection{Scenario 1: Scientific Data (End-to-End Validation)}

This scenario was designed to test the full IDEA2 workflow, encompassing the complete lifecycle from initial CQ extraction to iterative refinement.
The objective was to engineer requirements for a new ontology based on the Analytical Information Markup Language (AnIML) -- an XML-based standard for encoding experimental data in analytical chemistry and biology.
The AnIML standard consists of a \texttt{Core} and a \texttt{Technique Definitions} schemata. The former defines the structural framework for samples, experimental steps, and results, whereas the latter supports the annotation of Technique Definition Documents (ATDDs) that constrain this framework for specific analytical methods \cite{schaefer20XXschema}.

In the absence of explicit requirement specifications for AnIML, we utilise IDEA2 to elicit CQs directly from the \texttt{core} schema and selected technique definitions.
This approach aims to retrofit ontology requirements that are consistent with the legacy standard while bridging the gap to modern OE methodologies.
This scenario represents a compelling stress test for the workflow due to the complexity of the source material; the \texttt{core} document alone comprises approximately 2,500 lines of XML specification, requiring deep domain expertise to interpret them correctly. 
The evaluation engaged a panel of $N=4$ domain experts from our industry partner, Unilever PLC, with experience in using the AnIML standard ranging from 2 to 10 years.
SPARQL queries were used to further validate the final ontology against the validated CQs from IDEA2 \cite{morlidge2025animowl}.

\subsection{Scenario 2: Cultural Heritage (Reformulation Stress Test)}

The second scenario has been developed to specifically isolate and validate the \textbf{reformulation performance} of the workflow.
Rather than generating requirements \emph{ab initio}, this experiment focuses on repairing a set of previously rejected CQs from the an existing publicly available dataset (\emph{AskCQ} \cite{alharbi2025comparative}). For this experiment, the requirements were originally elicited from a user story involving a music archivist and a collection curator, focusing on their interactions with a museum's collection management system.

For this validation, we select the subset of 23 CQs that had received negative consensus scores (i.e., were rejected) in the original study.
To ensure a balanced assessment, this subset comprises a mix of:
\begin{enumerate*}[label=(\roman*)]
    \item human-authored CQs (11 items from subsets \textit{HA-1, HA-2} of two annotators); and
    \item LLM-generated CQs (12 items from subsets \textit{GPT, Gemini}) produced via zero-shot prompting of two different LLMs.
\end{enumerate*}
This balanced composition ensures that the evaluation of the reformulation engine is not biased towards a specific origin of the requirement (i.e., human vs. machine).

The rejected CQs were fed into the IDEA2 reformulation loop, enriched with the anonymised feedback (scores, comments) provided by the original annotators.
To enable informed validation, the dashboard displays a ``Recap'' field for each item, presenting the expert panel with the original CQ text, its rejection statistics, and the historical feedback that triggered the reformulation.
The resulting requirements were then reviewed by a panel of $N=3$ experts in the CH domain to determine if the refinement successfully resolved the original issues.

\begin{table}[t]
\centering
\caption{Comparative summary of key metrics from validation scenarios. Note: Final Acceptance Rate is calculated as the total number of accepted requirements divided by the total number of validation tasks (initial + reformulations).}
\label{tab:results}
\begin{tabular}{@{}lll@{}}
\toprule
\textbf{Metric} & \textbf{Scenario 1: AnIML} & \textbf{Scenario 2: CH} \\ \midrule
\textbf{Focus} & Full Workflow & CQ Reformulation \\
\textbf{Source Type} & XML Schema & Personas, User Story \\
\textbf{No. of Experts} & 4 & 3 \\
\textbf{Initial CQs} & 103 (IDEA2-generated) & 23 (Manual+LLM CQs) \\
\textbf{Iterations} & 3 & 2 \\
\textbf{Final Acceptance Rate} & 92.7\% (102/110) & 73.3\% (22/30) \\
\textbf{Avg. Time/Iteration} & 10.78 min & 14.85 min \\
\textbf{Avg. Time per CQ} & 0.85 min & 2.97 min \\ \bottomrule
\end{tabular}
\end{table}
\begin{table}[t]
\centering
\caption{Detailed breakdown of validation metrics per iteration, showing the number of CQs reviewed in each cycle. The decreased volume in later iterations corresponds to the subset of requirements requiring reformulation.}
\label{tab:iteration-metrics}
\begin{tabular}{@{}llccc@{}}
\toprule
\textbf{Scenario} & \textbf{Iteration} & \textbf{\# CQs} & \textbf{Avg. Time per CQ} & \textbf{Acceptance Rate} \\ \midrule
\multirow{3}{*}{\textbf{AnIML}} & It. 1 (Initial) & 103 & 0.58 min & 95.1\% \\
 & It. 2 (Reformulation) & 5 & 2.93 min & 60.0\% \\
 & It. 3 (Reformulation) & 2 & 1.75 min & 50.0\% \\ \midrule
\multirow{2}{*}{\textbf{CH}} & It. 1 (Reformulation) & 23 & 2.26 min & 69.6\% \\
 & It. 2 (Reformulation) & 7 & 3.19 min & 85.7\% \\ \bottomrule
\end{tabular}
\end{table}
\begin{table}[ht]
\centering
\caption{Comparison of inter-annotator agreement metrics across the two studies.}
\label{tab:agreement-comparison}
\begin{tabular}{lcc}
\hline
\textbf{Metric} & \textbf{AnIML (N=4)} & \textbf{CH (N=3))} \\
\hline
Observed agreement (\%) & 89.8 & 90.0 \\
Krippendorff's $\alpha$ \cite{krippendorff2004measuring} & 0.177 & 0.843 \\
Fleiss' $\kappa$ \cite{fleiss1971measuring} & 0.176 & 0.841 \\
Gwet's AC1 \cite{gwet2008computing} & 0.884 & 0.885 \\
PABAK \cite{byrt1993bias} & 0.797 & 0.867 \\
Pairwise Cohen's $\kappa$ \cite{cohen1960coefficient} range & 0.017 to 0.422 & 0.772 to 1.000 \\
\hline
\end{tabular}
\end{table}

\subsection{Results}
We evaluate IDEA2 across the two real-world scenarios in Section~\ref{sec:experiments} to assess its efficiency in eliciting requirements and the effectiveness of its iterative reformulation loop. First, we analyse the level of agreement between annotators in both scenarios, summarising key statistics and results in Table~\ref{tab:results}. Next, we report on the progression of acceptance rates and time costs across iterations in Table~\ref{tab:iteration-metrics} and we address potential bias in agreement formation in Table~\ref{tab:agreement-comparison}. The experiments behind these results are described in the remainder of this section.

\begin{figure}[t]
    \centering
    \begin{subfigure}[b]{0.95\textwidth}
        \centering
        \includegraphics[width=\linewidth]{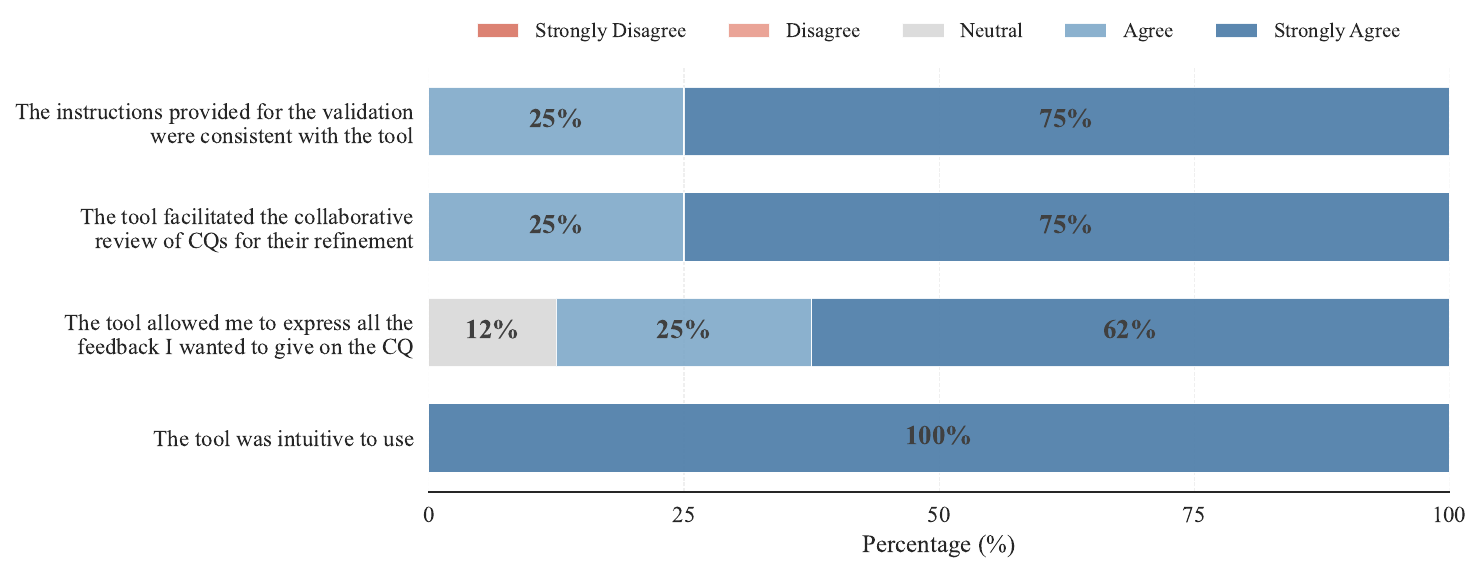}
        \caption{Scenario 1: AnIML}
        \label{fig:animl-usage}
    \end{subfigure}
    \par\bigskip 
    \begin{subfigure}[b]{0.95\textwidth}
        \centering
        \includegraphics[width=\linewidth]{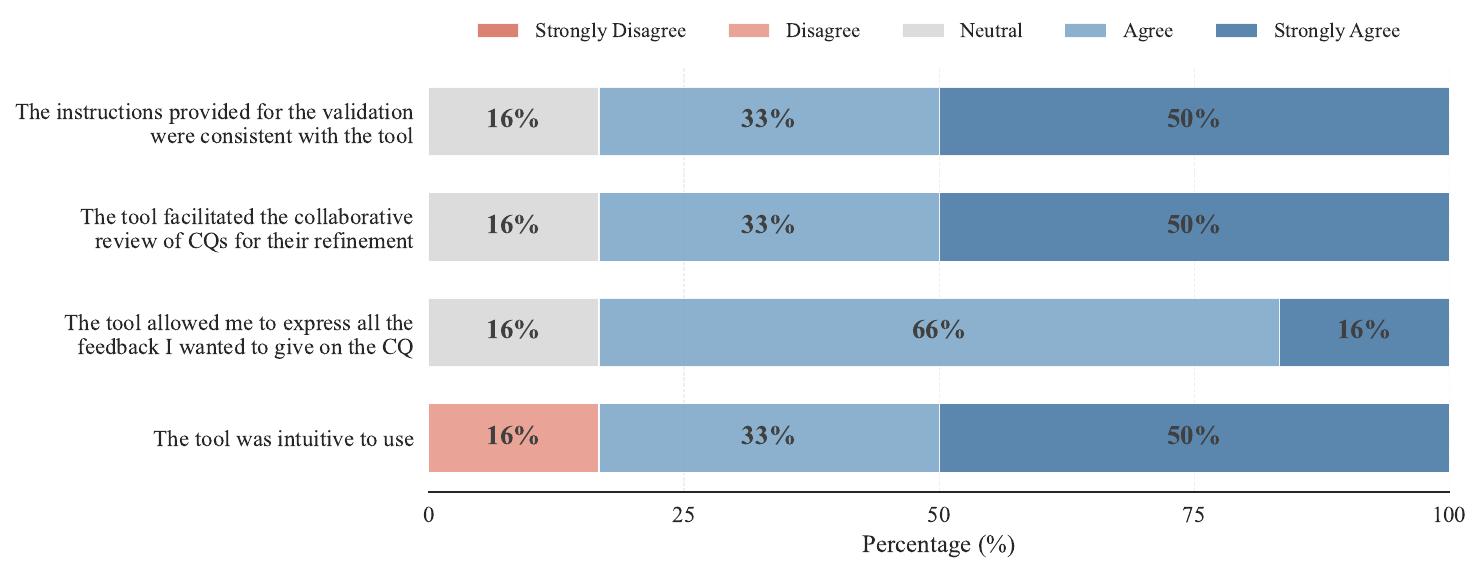}
        \caption{Scenario 2: Cultural Heritage}
        \label{fig:bme-usage}
    \end{subfigure}
    \caption{Usage metrics results derived from the expert evaluation. The Likert scale charts illustrate the domain experts' agreement regarding the tool's consistency, collaborative capabilities, expressiveness, and intuitiveness across (a) the AnIML scenario ($N=4$) and (b) the Cultural Heritage scenario ($N=3$).}
    \label{fig:usage-metrics}
\end{figure}

\subsubsection{Scenario 1: Scientific Data (AnIML).}
In the AnIML scenario, the initial extraction produced 103 candidate CQs derived from the XML schema. The first review iteration took an average of 23.33 minutes per expert, achieving a high initial acceptance rate of 95.1\% (98 out of 103 CQs). The 5 rejected CQs were automatically processed by the reformulation loop based on the experts' feedback.
After the first reformulation loop (Iteration 2), which took an average of 6.00 minutes, 3 additional CQs were accepted (60\% acceptance), demonstrating rapid convergence. A final, minor iteration (Iteration 3) was required for the remaining 2 CQs, taking 3.00 minutes and resulting in 1 final acceptance. In total, 102 CQs were accepted across 3 iterations from 110 review events (initial + reformulations), yielding a final cumulative acceptance rate of 92.7\%.

\subsubsection{Scenario 2: Cultural Heritage (CH).}
The second scenario focuses specifically on the effectiveness of the reformulation mechanism using requirements derived from personas and user stories. The initial iteration involved 23 CQs and took an average of 20.70 minutes. In contrast to the schema-driven AnIML scenario, the initial acceptance rate here was lower at 69.6\% (16 out of 23), reflecting the increased difficulty in reformulating CQs with little or no feedback.

However, the reformulation loop proved highly effective after the evaluators provided comments on the reformulations.
In the second iteration, which took an average of 9.00 minutes, 6 out of the 7 rejected CQs were successfully reformulated and accepted (85.7\% acceptance). This significant improvement in the second pass confirms that the expert-in-the-loop mechanism successfully captures the necessary context to resolve ambiguities in narrative requirements.

\paragraph{Inter-Annotator Agreement Analysis}
We assess reliability among four annotators for the AnIML CQs, using both raw and chance-corrected metrics.
The observed agreement was high at 89.8\%, indicating that annotators often accepted the same CQs.
However, chance-corrected measures revealed lower reliability due to pronounced category imbalance (93.4\% of ratings were ``1'', denoting \textit{acceptance}): Krippendorff's $\alpha$ was 0.177 and Fleiss' $\kappa$ was 0.176.
To address the prevalence effect, we additionally compute Gwet's AC1 and PABAK, which assess inter-rater agreement while adjusting for chance, with Gwet’s AC1 providing stability in cases of category imbalance and PABAK correcting for both prevalence and bias effects. Gwet's AC1 yielded 0.884, and PABAK was 0.797, both substantially higher than $\alpha$ and $\kappa$, reflecting that most disagreements occurred in the minority class.
Pairwise Cohen's $\kappa$ ranged from 0.017 to 0.422, indicating heterogeneous agreement between individual annotators.
These findings underscore that while raw agreement appears strong, reliability beyond chance is modest under extreme prevalence, and complementary metrics such as AC1 and PABAK provide a more nuanced view of annotation consistency.

In comparison, for the Cultural Heritage (CH) scenario, the observed agreement was 90.0\%, and pairwise Cohen's $\kappa$ ranged from 0.772 (Annotator 1-2 and 2-3) to 1.000 (Annotators 1-3), indicating strong consistency between annotators.
Chance-corrected measures confirm high reliability: Krippendorff's $\alpha$ was 0.843 and Fleiss' $\kappa$ was 0.841. To account for prevalence effects (70.0\% of ratings were ``1'', denoting CQ acceptance), we compute Gwet's AC1 and PABAK, which yielded 0.885 and 0.867 respectively, both supporting substantial agreement, and higher than in the previous scenario.

\subsubsection{Usage metrics and usability.}
To assess the usability of the collaborative workflow, participants completed a post-task survey comprising four Likert-scale questions regarding instruction clarity, collaboration facilitation, expressiveness, and intuitiveness. The feedback was predominantly positive across both domains.

In the AnIML scenario (Fig.~\ref{fig:usage-metrics}a), the results indicate high user acceptance, with all the domain experts strongly agreeing that the tool was intuitive to use.
Furthermore, all participants agreed or strongly agreed that the tool effectively facilitated the collaborative review process. 
One participant also mentioned that this process expanded their knowledge of the AnIML schema by discovering new features that were previously unknown through the validation of CQs.

The Cultural Heritage scenario (Fig.~\ref{fig:usage-metrics}b) mirrored these positive trends, particularly regarding the tool's expressiveness, where over 80\% of responses were positive (Agree or Strongly Agree).
While the results for this group exhibited slightly higher variance, including a minority of neutral or negative responses regarding intuitiveness, the overall consensus suggests that the Notion dashboard provides an accessible platform for requirements engineering.

\section{Discussion}\label{sec:discussion}

The experimental results highlight several key aspects of the IDEA2 workflow's design and its broader implications for requirements engineering practice.

The evaluation’s key finding is the effectiveness of the iterative reformulation loop, which transforms unstructured human insight into structured, machine-actionable input. Simply asking an LLM to generate or improve questions is ineffective \cite{alharbi2025comparative}, but providing a rejected question with specific reasons and expert feedback enables targeted revisions. The AnIML scenario illustrates this: an ambiguous CQ -- ``What result blueprints are defined for a technique?'' -- was rejected, then refined via expert input to ``What are the names of the result definitions specified by a technique?'' and accepted with a score of 3.
The revised version was easily understood and accepted, demonstrating the workflow's capacity to act as a mediator, resolving terminological ambiguities that often arise between technical documentation and the conceptual understanding of DEs.

The IDEA2 workflow suggests a potential evolution in the roles and dynamics of the requirements engineering process.
Traditionally, the knowledge engineer acts as a primary interviewer and interpreter, translating conversations with domain experts into formal specifications.
This workflow reframes these roles, and the knowledge engineer transitions from interviewer to facilitator in a hybrid human-AI collaborative system. Their primary tasks become setting up the workflow, designing initial extraction prompts, and monitoring the process.
The DE's role is elevated from passive interviewee to active co-creator and validator.
They are given direct agency to shape, approve, and reject requirements within the system, making them central and empowered participants in the design process.
This dynamic aligns with modern agile development that emphasises continuous stakeholder collaboration and direct feedback loops \cite{cao2008agile}.

Despite the promising results, this study has limitations that inform avenues for future work.
The effectiveness of the reformulation loop is critically dependent on the quality and richness of the feedback provided by experts.
This is a well-documented challenge in human-in-the-loop systems, where low-quality or inconsistent human input can degrade rather than improve model performance \cite{wu2022survey}.
Future work should investigate how to guide experts towards providing more actionable feedback.
Furthermore, the current workflow relies on consensus via majority vote, which may oversimplify complex situations. Expert disagreement can signal genuine ambiguity in requirements or multiple valid perspectives, rather than mere error \cite{aroyo2015crowdtruth}. Rather than enforcing consensus, future iterations could adopt crowdsourcing and peer-review techniques to better manage disagreement. This might include flagging contentious CQs for moderated discussion or weighting votes by annotator confidence and expertise, thereby preserving important minority viewpoints \cite{dumitrache2017crowdtruth}.

Sustaining a high level of expert engagement throughout a lengthy requirements engineering process is another significant challenge. The risk of ``annotator fatigue'' can lead to a drop in feedback quality over time, particularly in large-scale projects (100+ CQs).
To mitigate this, future versions could implement strategies to maintain engagement, such as presenting CQs in smaller, thematically organised batches.
Moreover, the integration of gamification elements -- such as points, badges, or progress tracking -- could be explored to incentivise continued and high-quality participation, a technique that has shown promise in other areas of software engineering \cite{mutsuddi2024cocreating}.
Finally, while the use of Notion proved effective for rapid prototyping and demonstrating the workflow's viability, implementing these more advanced features for managing disagreement and engagement would likely necessitate a more customised platform.

\section{Conclusions}\label{sec:conclusions}

This paper introduced IDEA2, a semi-automated workflow designed to overcome the knowledge acquisition bottleneck in ontology engineering.
By decoupling requirement elicitation (delegated to LLMs) from requirement validation (reserved for domain experts), and mediating their interaction through a collaborative feedback loop, IDEA2 streamlines the authoring of competency questions.

Our empirical evaluation across two heterogeneous domains -- scientific data standards and cultural heritage -- confirmed the effectiveness of this hybrid approach.
In the end-to-end \textit{AnIML} scenario, the workflow demonstrated high efficiency, achieving a cumulative acceptance rate of 92.7\% with rapid convergence across iterations.
Crucially, the \textit{Cultural Heritage} stress test validated the core reformulation mechanism: the system successfully repaired 85.7\% of previously rejected requirements by leveraging expert critique, proving that LLMs can effectively interpret and act upon structured human feedback.
Furthermore, the usability assessment indicates that integrating the workflow into a familiar, non-specialist platform (Notion) significantly lowers the technical barrier to entry, fostering high levels of engagement and expressiveness among domain experts.

Future research will focus on enhancing the robustness of the expert-in-the-loop mechanism.
First, as the reformulation quality is contingent on the richness of human input, we aim to integrate NLP techniques to assess expert comments in real-time, prompting for clarification if feedback is too generic.
Second, we plan to move beyond simple majority voting by adopting ``disagreement-aware'' consensus models (e.g., CrowdTruth), treating expert divergence not as noise, but as a signal of genuine domain ambiguity to be preserved.

\section*{Acknowledgements}
This work was supported by the Royal Academy of Engineering under the Google DeepMind Research Ready Scheme.

\bibliographystyle{plain}
\bibliography{references}

\end{document}